%% file: SimVA.tex
\documentclass[lettersize,journal]{IEEEtran}
\usepackage{amsmath,amsfonts}
\usepackage{algorithmic}
\usepackage{algorithm}
\usepackage{array}
\usepackage[caption=false,font=normalsize,labelfont=sf,textfont=sf]{subfig}
\usepackage{textcomp}
\usepackage{stfloats}
\usepackage{url}
\usepackage{verbatim}
\usepackage{graphicx}
\usepackage{cite}
\usepackage{multirow}
\usepackage{makecell}
\usepackage{soul}      
\usepackage{fontawesome5}
\usepackage[table,dvipsnames]{xcolor} 

\colorlet{colorFst}{Green!25}
\colorlet{colorSnd}{SpringGreen!45} 

\definecolor{yrcolor}{rgb}{0.15,0.35,0.75}

\definecolor{jycolor}{rgb}{1,0,1}

\definecolor{dbcolor}{rgb}{1,0,0}

\definecolor{sjcolor}{rgb}{0.00,0.50,0.48}

\begin{document}

\title{Spatio-Temporal Similarity Volume Aggregation\\ for Open-Vocabulary Action Recognition}

\author{Yerim So, Jiyeong Kim, Jiwon Yoon, Sangryul Jeon, Dongbo Min
\thanks{Yerim So, Jiyeong Kim, Jiwon Yoon and Dongbo Min are with the Department of Computer Science and Engineering, Ewha Womans University, Seoul, South Korea (e-mail: yrso@ewha.ac.kr; wldud8946@ewha.ac.kr; jiwon105020@ewha.ac.kr; dbmin@ewha.ac.kr).}
\thanks{Sangryul Jeon is with the Department of Computer Science and Engineering, Pusan National University, Busan, South Korea (e-mail: srjeonn@pusan.ac.kr).}
}

\markboth{IEEE Transactions on Image Processing}
{So \MakeLowercase{\textit{et al.}}: Spatio-Temporal Similarity Volume Aggregation}


\maketitle

\input{section/0_abstract}

\begin{IEEEkeywords}
Open-Vocabulary Action Recognition, Spatio-Temporal Similarity Volume
\end{IEEEkeywords}

\input{section/1_intro}
\input{section/2_method}
\input{section/3_exp1}
\input{section/4_related}
\input{section/5_conclusion}



{\small
\bibliographystyle{IEEEtran}
\bibliography{section/9_refs}
}

\section{Biography Section}

\begin{IEEEbiography}[{\includegraphics[
    width=1in,
    height=1.25in,
    clip,
    keepaspectratio
]{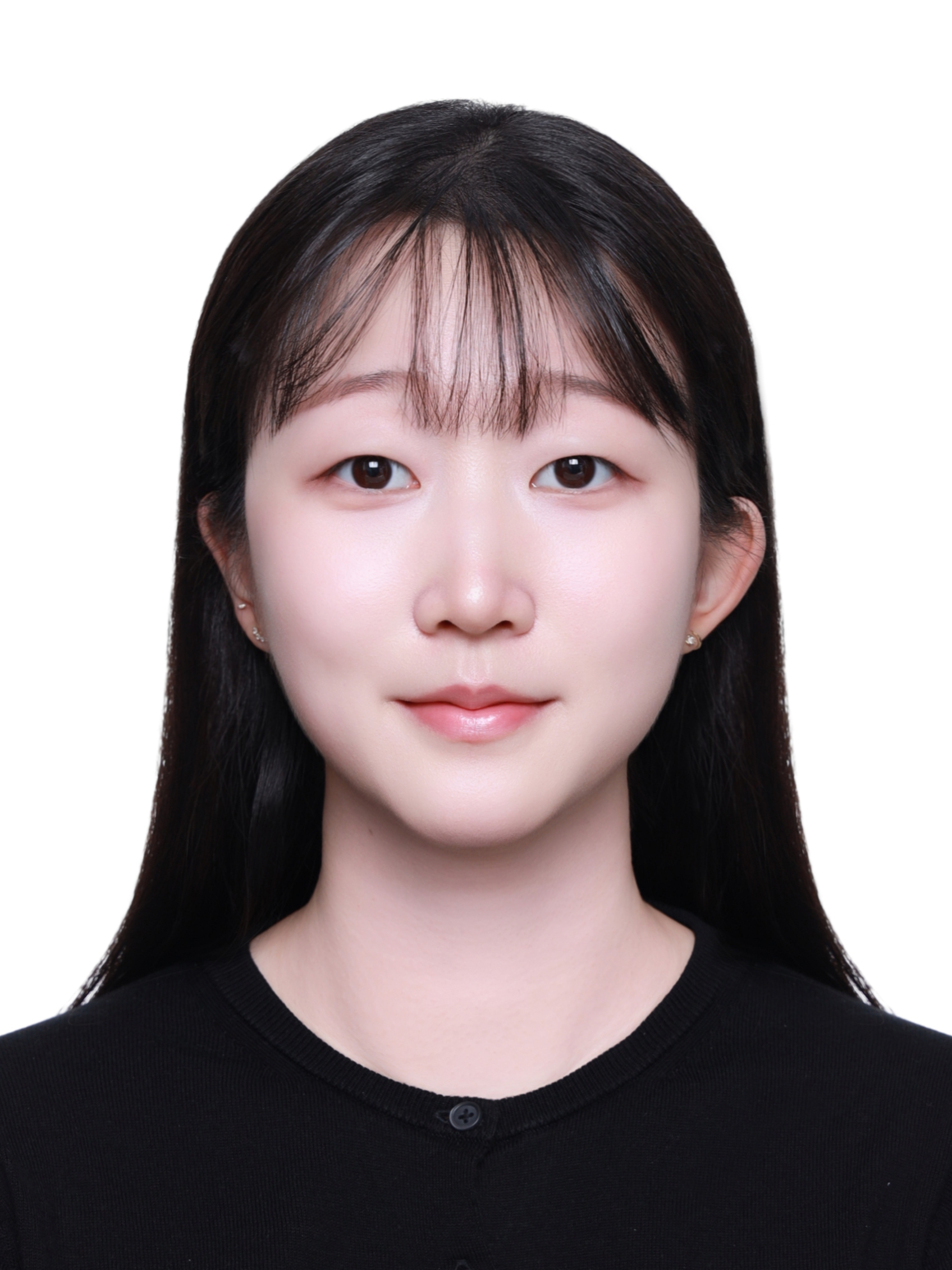}}]{Yerim So}
has been pursuing the M.S. degree at the School of Computer Science and Engineering, Ewha Womans University, Seoul, South Korea, since 2025.
Her research interests include video understanding and multimodal learning.
\end{IEEEbiography}

\begin{IEEEbiography}[{\includegraphics[
    width=1in,
    height=1.25in,
    clip,
    keepaspectratio
]{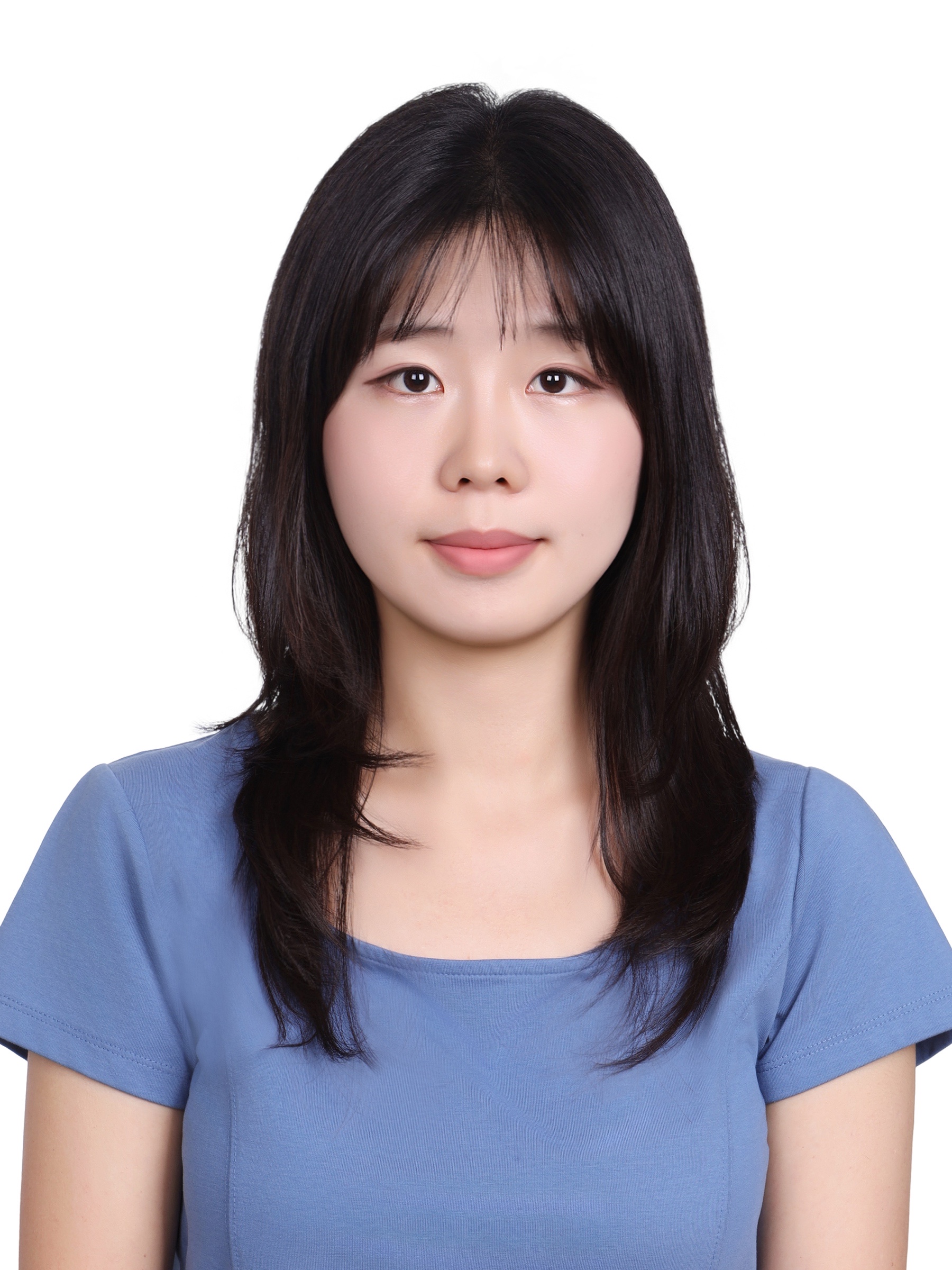}}]{Jiyeong Kim} is currently pursuing joint M.S. and Ph.D. degrees in the Department of Computer Science and Engineering, Ewha Womans University, Seoul, South Korea. Her current research interests include computer vision, unified multimodal models, model compression, and video understanding.
\end{IEEEbiography}

\begin{IEEEbiography}[{\includegraphics[
    width=1in,
    height=1.25in,
    clip,
    keepaspectratio
]{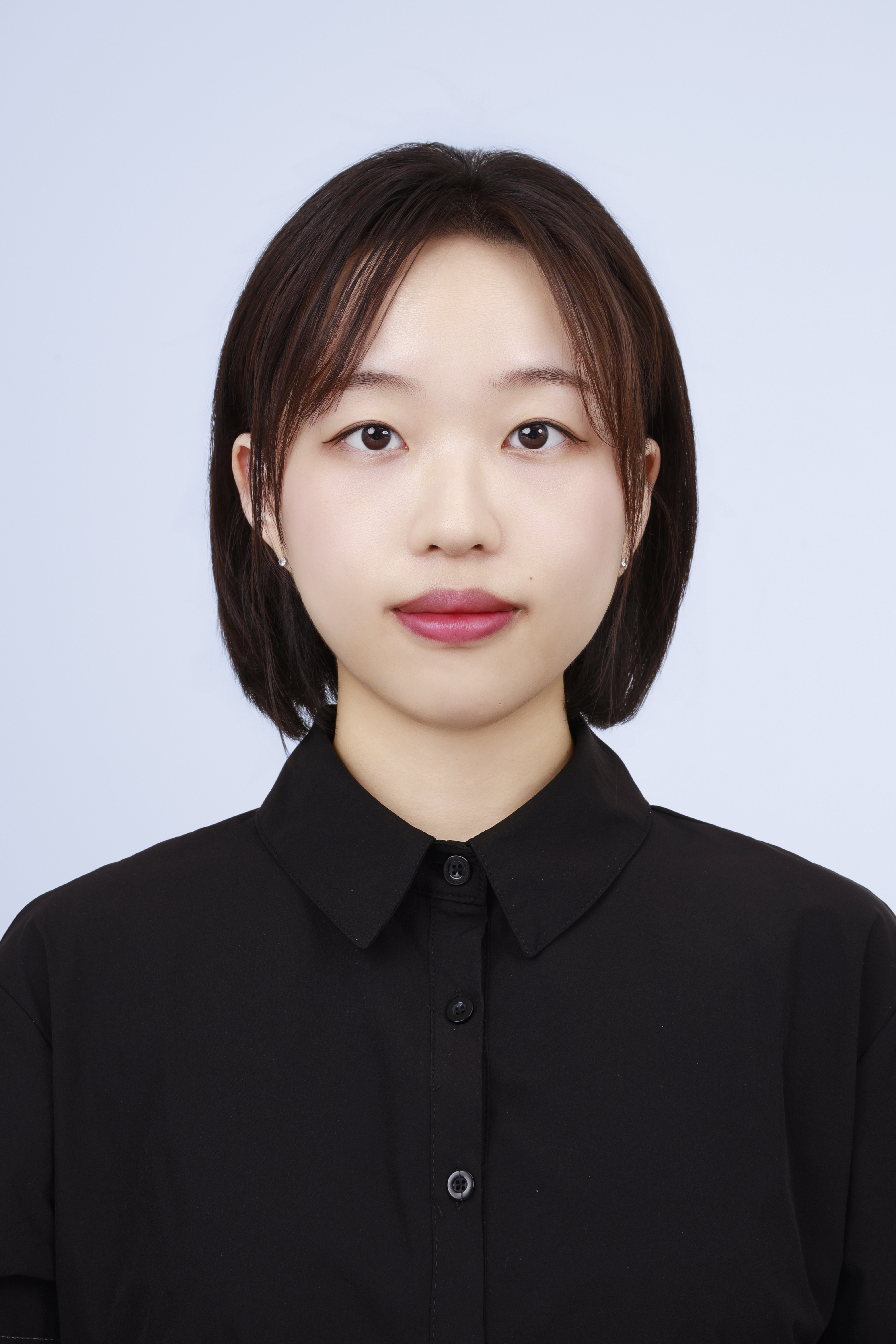}}]{Jiwon Yoon} received the M.S. degree from the School of Computer Science and Engineering, Ewha Womans University, Seoul, South Korea, in 2026. She has been a Research Engineer with LG AI Research, Seoul, South Korea, since 2026. Her research interests include computer vision, multimodal learning, and bioinformatics.
\end{IEEEbiography}

\begin{IEEEbiography}[{\includegraphics[
    width=1in,
    height=1.25in,
    clip,
    keepaspectratio
]{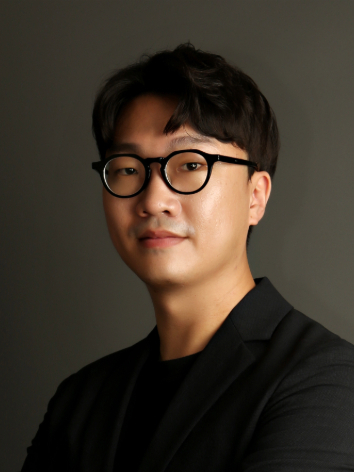}}]{Sangryul Jeon} received the B.S. and Ph.D. degrees in Electrical and Electronic Engineering from Yonsei University, Seoul, South Korea, in 2016 and 2022, respectively. From 2022 to 2023, he was a Postdoctoral Researcher with the Department of Electrical Engineering and Computer Sciences at the University of California, Berkeley. Since 2023, he has been an Assistant Professor with the Department of Computer Science and Engineering at Pusan National University, Busan, South Korea. His research interests include 2D/3D computer vision and machine learning, with a particular focus on image–text alignment and representation learning.
\end{IEEEbiography}

\begin{IEEEbiography}[{\includegraphics[
    width=1in,
    height=1.25in,
    clip,
    keepaspectratio
]{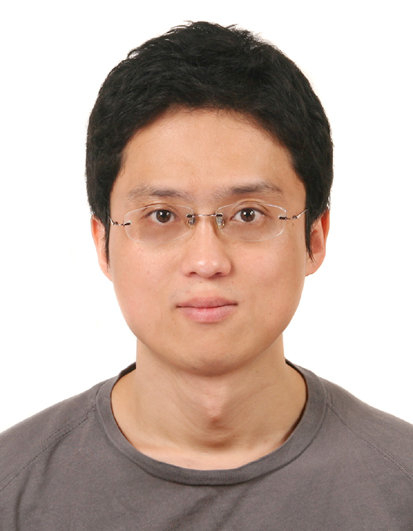}}]{Dongbo Min} (Senior Member, IEEE) received the B.S., M.S., and Ph.D. degrees from the School of Electrical and Electronic Engineering, Yonsei University, Seoul, South Korea, in 2003, 2005, and 2009, respectively. From 2009 to 2010, he was a Postdoctoral Researcher at Mitsubishi Electric Research Laboratories, Cambridge, MA, USA. From 2010 to 2015, he was at
Advanced Digital Sciences Center, Singapore. From 2015 to 2018, he was an
Assistant Professor at the Department of Computer Science and Engineering,
Chungnam National University, Daejeon, South Korea. Since 2018, he has
been working with the Department of Computer Science and Engineering,
Ewha Womans University, Seoul, South Korea. His current research interests
include computer vision, deep learning, and video processing.
\end{IEEEbiography}

\end{document}

%% file: section/0_abstract.tex
\begin{abstract}
Recent Open-Vocabulary Action Recognition (OVAR) methods typically aggregate visual features into a global representation before computing text alignment, 
a process that can obscure local patch information and fine-grained spatio-temporal cues.
We propose \textbf{Sim}ilarity \textbf{V}olume \textbf{A}ggregation (SimVA), a framework that constructs a dense 4D spatio-temporal similarity volume from patch-level visual-text similarities. 

The similarity volume is refined by spatial aggregation, which contextualizes local similarity patterns to improve intra-frame consistency.
To mitigate the computational overhead of aggregating the high-dimensional similarity volume over large vocabularies, we employ class sampling during training while retaining all candidate classes at inference.
Motion-aware modulation incorporates variation cues derived from consecutive frame features.
Mamba-based temporal aggregation then models the evolution of class-conditioned similarity patterns across frames. By maintaining dense visual-text correspondence, SimVA effectively transfers CLIP to video action recognition, achieving competitive performance across zero-shot, few-shot, and base-to-novel benchmarks.
\end{abstract}

%% file: section/1_intro.tex
\section{Introduction}

\IEEEPARstart{R}{ecent} advances in large-scale Vision-Language Models (VLMs)~\cite{align, blip, siglip, siglip2} have led to substantial progress in open-vocabulary visual recognition. By learning a joint embedding space through contrastive training on visual–text pairs, CLIP~\cite{CLIP} acquires robust and transferable visual representations that exhibit strong zero-shot generalization to unseen categories. This capability has rapidly been extended beyond image classification into pixel- and region-level prediction tasks, including semantic segmentation~\cite{OpenSeg, OVSeg, Catseg} and object detection~\cite{RegionCLIP, Detic, GLIP}. Notably, the characteristics of VLMs offer a promising avenue for mitigating the severe data acquisition and annotation challenges inherent in video understanding~\cite{howto100m, videoclip}. Since collecting and annotating frame-level video data is substantially more expensive than image data, adapting pre-trained image-based VLMs to the video domain without requiring additional large-scale video–text pre-training has emerged as a practical and effective research direction~\cite{Actionclip, xclip, vificlip, tcclip, bdcclip}.

\begin{figure}[t]
  \centering
  \includegraphics[width=\linewidth]{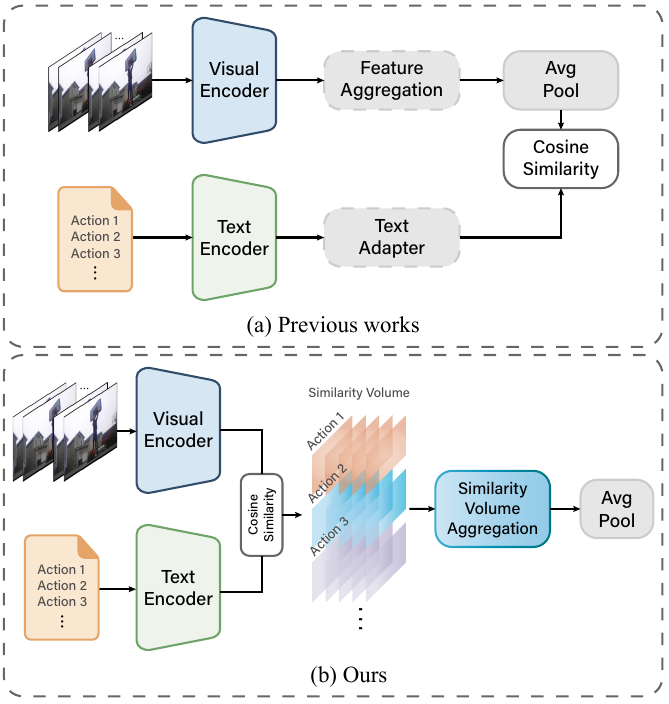}
  \caption{\textbf{Conceptual comparison of aggregation paradigms for Open-Vocabulary Action Recognition.} (a) Representative feature-aggregation methods~\cite{Actionclip, xclip, tcclip, bdcclip} follow an aggregate-then-align paradigm: visual features undergo class-agnostic spatio-temporal aggregation before the resulting video-level representation is compared with candidate action text embeddings. (b) SimVA follows an align-then-aggregate paradigm: patch-level visual tokens are first compared with candidate action text embeddings to construct a class-conditioned spatio-temporal similarity volume. Spatial and temporal aggregation then refine the matching evidence for each candidate action before video-level pooling.
  }
  \label{fig:1}
\end{figure}

To adapt image-based VLMs to the video domain, Open-Vocabulary Action Recognition (OVAR) methods typically follow an \emph{aggregate-then-align} paradigm (Fig.~\ref{fig:1}(a)), in which visual features are first aggregated into a video-level representation that is subsequently compared with candidate action text embeddings. For example, ActionCLIP~\cite{Actionclip} models inter-frame relationships using a temporal Transformer and then compares the resulting video-level representation with candidate action text embeddings. ViFi-CLIP~\cite{vificlip} constructs a video-level representation by average pooling frame-level visual features before computing its similarity to candidate action text embeddings.

\IEEEpubidadjcol Recent approaches have sought to exploit local information more effectively within this framework. TC-CLIP~\cite{tcclip} aggregates informative visual tokens across frames into context tokens, and BDC-CLIP~\cite{bdcclip} models correlations among visual tokens via Brownian Distance Covariance before video--text alignment. These advances highlight the importance of fine-grained visual representations. However, spatio-temporal aggregation of local tokens remains independent of candidate action semantics. Consequently, the aggregated representation may emphasize appearance cues common to multiple action categories while insufficiently capturing action-discriminative spatial relations and temporal dynamics.

\begin{figure*}[t]
  \centering
  \includegraphics[width=\textwidth]{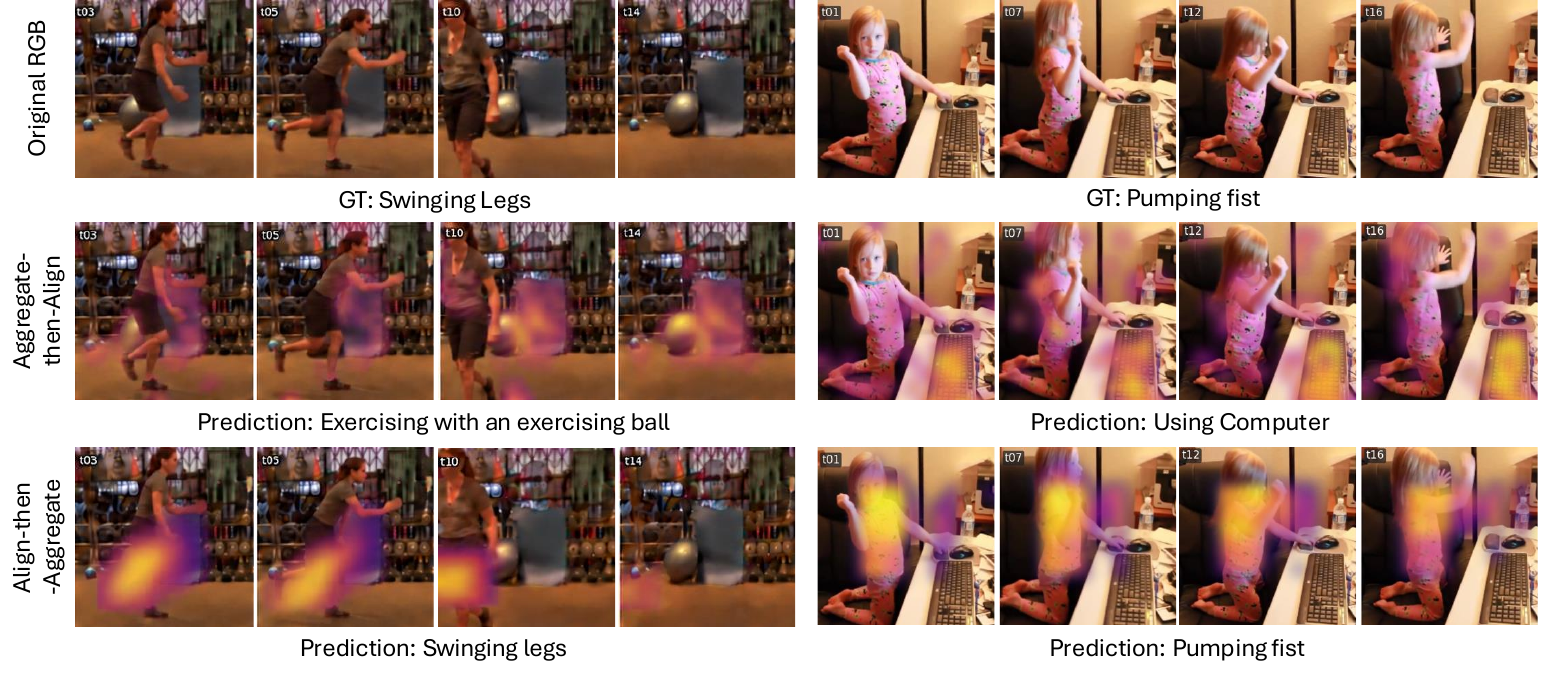}
  \caption{\textbf{Qualitative comparison of aggregate-then-align and align-then-aggregate.} Rows show input RGB frames (top), activation maps and predictions from aggregate-then-align (middle), and those from SimVA (bottom). SimVA exhibits more focused activations around action-relevant body regions and correctly recognizes Swinging legs (left) and Pumping fist (right), both of which are misclassified by aggregate-then-align. GT denotes the ground-truth action class. Additional quantitative results are reported in Tab.~\ref{tab:motion}.}
  \label{fig:5}
\end{figure*}

This limitation can be particularly pronounced when surrounding objects or scene context suggest an action different from the one being performed. For example, in Fig.~\ref{fig:5}, the \emph{aggregate-then-align} model misclassifies \textit{Swinging Legs} as \textit{Exercising with an Exercising Ball} and \textit{Pumping Fist} as \textit{Using Computer}. The ball and gym setting in the first video and the computer and desk in the second provide contextual cues associated with their respective incorrect classes. However, the key evidence for identifying the actual actions lies in leg postures and changes in leg position across frames for \textit{Swinging Legs}, and in the movement of the raised arm and fist for \textit{Pumping Fist}. If such local discriminative evidence is not adequately preserved during aggregation, subsequent video--text alignment may struggle to exploit it, as it operates on an already compressed representation. This motivates an aggregation process conditioned on candidate action semantics, in which local evidence supporting each candidate is refined across space and time before being integrated into a video-level prediction.

To address this limitation, we propose \textbf{Sim}ilarity \textbf{V}olume \textbf{A}ggregation (\textbf{SimVA}), which incorporates candidate action semantics into spatio-temporal aggregation through an \emph{align-then-aggregate} paradigm (Fig.~\ref{fig:1}(b)). Instead of first compressing visual features into a video-level representation, SimVA computes patch-level similarities between visual tokens and candidate action text embeddings and organizes these responses into a dense spatio-temporal similarity volume. This preserves where and when the video supports each candidate, allowing aggregation to refine candidate-specific matching evidence before producing video-level scores. Fig.~\ref{fig:5} presents qualitative examples in which SimVA exhibits more focused activations in class-relevant regions and makes correct predictions. Because the aggregation modules share their parameters across candidates, the same process can be applied to matching responses computed from unseen action text embeddings. During training, we select high-scoring candidate actions based on global video--text similarity while maintaining stochastic exploration of additional classes. Candidates with high global similarity may share coarse appearance or scene context with the ground-truth action, so training on these alternatives encourages discrimination based on finer-grained spatial and temporal matching patterns.

To extract coherent action semantics, we progressively refine the similarity volume through spatial aggregation followed by temporal aggregation. Spatial aggregation models relationships among neighboring matching responses to improve intra-frame consistency, since an individual response alone does not establish whether surrounding regions support the same action. However, this frame-wise refinement does not explicitly incorporate how visual regions change between frames. Motion-aware modulation introduces variation cues estimated from consecutive visual features to emphasize dynamically changing regions in the spatially refined evidence. These cues describe local transitions between adjacent frames and must be integrated over time to capture the temporal structure of an action. Mamba-based temporal aggregation performs sequential state updates over the motion-enhanced matching patterns, accumulating candidate-specific evidence in temporal order.

Our main contributions are three-fold:
\begin{itemize}
\item We propose SimVA, an align-then-aggregate framework for CLIP-based OVAR that shifts aggregation from visual features to class-conditioned similarity representations, enabling the direct refinement of dense spatio-temporal visual--text correspondences.
\item We design a structured similarity aggregation architecture that combines spatial contextualization, motion-aware modulation, and temporal modeling to refine candidate-specific matching evidence within and across frames.
\item We validate SimVA across zero-shot, few-shot, and base-to-novel action recognition benchmarks, achieving competitive performance and demonstrating generalization to unseen action categories.
\end{itemize}

%% file: section/2_method.tex
\begin{figure*}[t]
  \centering
  \includegraphics[width=\textwidth]{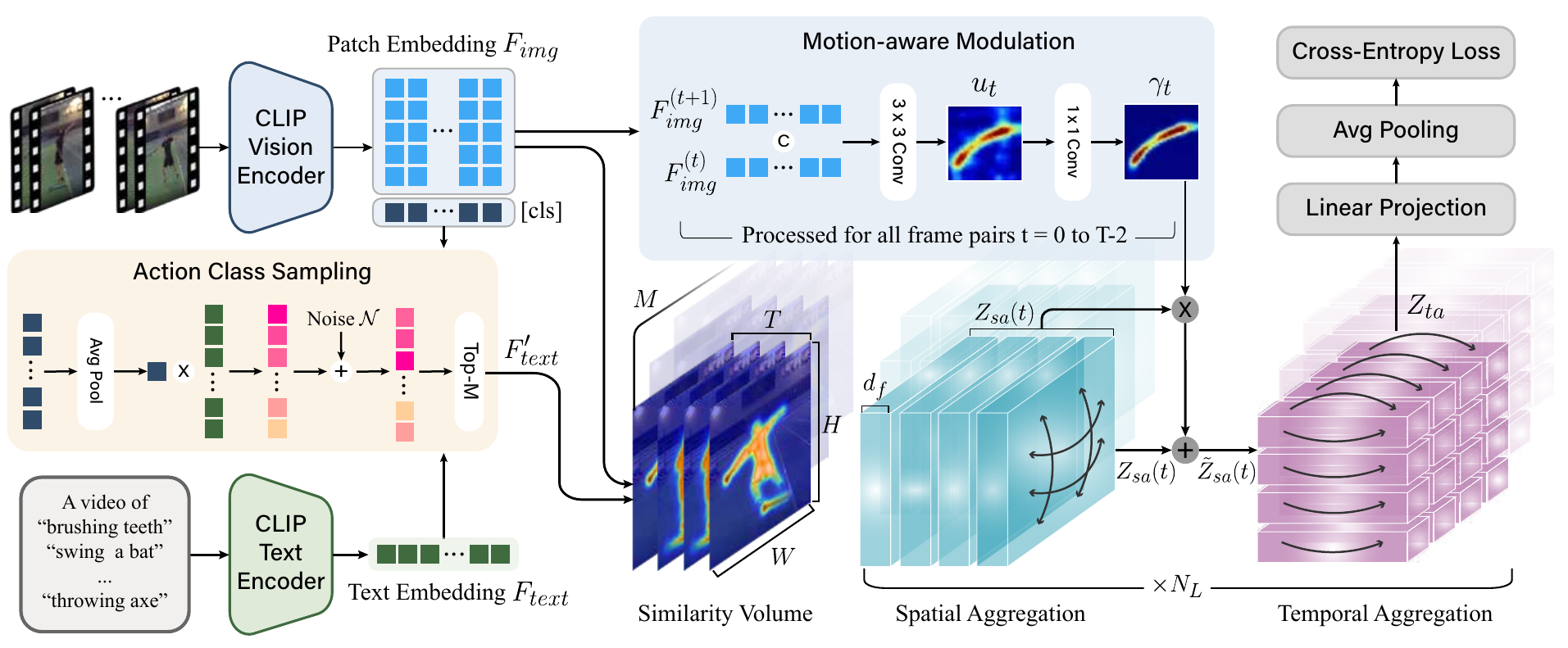}
  \caption{\textbf{Overall architecture of the SimVA framework.} Given an input video and action text prompts, we extract features using CLIP~\cite{CLIP}.
  During training, action class sampling (Sec.~\ref{sec:sampling}) selects a subset of candidate classes for similarity volume construction (~\ref{sec:volume}); at inference, the full evaluation vocabulary is used.
  This volume is then processed through a structured aggregation architecture: the spatial aggregation module (Sec.~\ref{sec:sa}) refines intra-frame context, motion-aware modulation injects adjacent-frame variation cues, and the temporal aggregation module (Sec.~\ref{sec:ta}) models the evolution of similarity patterns across frames.
  }
  \label{fig:2}
\end{figure*}

\section{Method}
\label{sec:method}

Fig.~\ref{fig:2} illustrates the overall architecture of the proposed SimVA framework.
Given an input video and a set of action class text prompts, SimVA extracts patch-level visual embeddings and text embeddings using the CLIP encoders.
Following an align-then-aggregate formulation, we first compute dense patch-level video-text similarities over action classes to construct a class-conditioned spatio-temporal similarity volume.
We introduce training-time action class sampling, which restricts the candidate vocabulary to a compact subset of visually relevant classes to encourage discrimination among potentially confusable actions, while incorporating stochastic exploration of additional classes.
The resulting class-conditioned similarity volume is refined through a structured aggregation pipeline: spatial aggregation contextualizes intra-frame similarity patterns, motion-aware modulation injects inter-frame variation cues, and temporal aggregation models the evolution of similarity patterns across frames.
Finally, the refined similarity representations are linearly projected into class scores and pooled over space and time to produce video-level action logits.
We describe each component in detail below.

\subsection{Similarity Volume Construction}
\label{sec:volume}
Aggregating patch-level evidence before cross-modal alignment can obscure fine-grained local cues that are important for grounding visual patterns to textual semantics~\cite{yaofilip,rao2022denseclip}.
To preserve such local evidence in videos, our framework constructs a dense similarity volume for fine-grained spatio-temporal alignment between video and text.
Given an input video $V \in \mathbb{R}^{T\times H_0\times W_0\times 3}$, we first extract patch-level embeddings $F_{img} = E_{img}(V)\in \mathbb{R}^{T \times H \times W \times D}$ of a spatial resolution $H \times W$ with channel dimension $D$ for $T$ frames using the CLIP image encoder $E_{img}$~\cite{CLIP}. 
Given action text prompts $L$, we construct text embeddings $F_{text}=E_{text}(L) \in \mathbb{R}^{N_C \times D}$ for action categories using the CLIP text encoder $E_{text}$, where $N_C$ denotes the total number of action classes. To evaluate the matching degree between each spatio-temporal location $(t, i, j)$ in the video and a specific class $c$, we compute the patch-wise cosine similarity to formulate a 4D spatio-temporal similarity volume $S \in \mathbb{R}^{T \times H \times W \times N_C}$:
\begin{align}
    S(t, i, j, c)=\frac{F_{img}(t, i, j, :)\cdot F_{text}(c, :)}{ \left\|F_{img}(t, i, j, :) \right\| \left\| F_{text}(c, :) \right\|}.
    \label{eq:c}
\end{align}
$S$ represents the similarity scores between all spatio-temporal locations of the video and each class.
We then apply a $7 \times 7$ convolution layer along the spatial dimension to project it into a $d_f$-dimensional space, $Z_0 = \text{Conv}_{7 \times 7}(S)$, where $\quad Z_0 \in \mathbb{R}^{T \times H \times W \times N_C \times d_f}$.

\subsection{Action Class Sampling}
\label{sec:sampling}
During training, we perform class sampling over the $N_C$ action classes to select a subset of candidate classes that are highly relevant to the input video. Visually similar candidates usually share coarse appearance cues with the ground-truth action, requiring finer-grained evidence to distinguish them. By focusing learning on these competing candidates, class sampling encourages the model to refine local video-text matching patterns that capture discriminative spatial and temporal cues beyond shared appearance information. This promotes fine-grained discrimination between the ground-truth action and visually similar alternatives.
Specifically, we compute a global video representation $\bar{V} \in \mathbb{R}^{D}$ by averaging the frame-level \texttt{[CLS]} tokens $v^t \in \mathbb{R}^D$ over time, followed by $\ell_2$ normalization:

\begin{align}
\bar{V} = \frac{v_{\mathrm{avg}}}{\|v_{\mathrm{avg}}\|}, \quad v_{\mathrm{avg}} = \frac{1}{T} \sum_{t=1}^T v^t.
\end{align}

We then compute prior similarity scores for all classes as $S_{global} = F_{text}\bar{V}$, and select a subset of the top-$M$ classes $\mathcal{I}_M$. During training, uniform noise $\mathcal{N} \sim \mathcal{U}([0, 0.5)^{N_C})$ is added to the scores to encourage exploration, and the ground-truth action class is always included in the sampled subset. Finally, the text embeddings of this selected subset, denoted as $F'_{text} \in \mathbb{R}^{M \times D}$, replace the original $F_{text}$ in Eq. \eqref{eq:c}. 
This strategy also reduces the computational burden during training, shrinking the class dimension of the similarity volume to a tractable size $M$, \textit{i.e.}, $Z_0 \in \mathbb{R}^{T \times H \times W \times M \times d_f}$. At inference, class sampling is omitted, and similarity volumes are constructed over all candidate action classes in the evaluation vocabulary.

\subsection{Spatio-Temporal Similarity Aggregation}
\label{sec:st_aggregation}
Given the sampled class-conditioned similarity volume $Z_0$, SimVA performs spatio-temporal aggregation directly in the similarity space. 
This preserves the candidate-class dimension throughout the aggregation process, allowing the model to refine the spatial distribution and temporal evolution of matching evidence for each candidate action.
This aggregation block is repeated for $N_L$ layers, with each layer sequentially applying spatial aggregation, motion-aware modulation, and temporal aggregation. After the final layer, a linear projection followed by spatio-temporal average pooling produces video-level action logits. We describe each operation below.

\subsubsection{Spatial Aggregation}
\label{sec:sa}
The constructed similarity volume contains class-specific responses at each spatio-temporal location. Spatial aggregation jointly refines these responses within each frame to incorporate the spatial structure of the matching patterns.
To address this, we apply spatial aggregation based on Swin Transformer~\cite{swin}. For a given time step $t$ and class $c$, we apply Swin Transformer blocks, denoted as $A_{sa}(\cdot)$,  featuring window-based self-attention and shifted window mechanisms to the spatial similarity embeddings of size $H \times W$:
\begin{equation}
    \begin{aligned}
        \label{eq:Zsa}
        Z_{sa}(t, :, :, c, :) = A_{sa}(Z_0(t, :, :, c, :)),
        \\
        \text{for}\;\; t=1,\dots,T, \;\; c\in \mathcal{I}_M.
    \end{aligned}
\end{equation}
By modeling intra-frame spatial interactions within each class-conditioned similarity map, spatial aggregation helps suppress noisy background responses and improve spatial consistency among patches belonging to the same object~\cite{Catseg, raft, cats, vat}. Consequently, this process yields more distinct local spatial patterns associated with the action.

\subsubsection{Temporal Aggregation}
\label{sec:ta}
Although spatial aggregation refines the similarity representations within each frame, the resulting features remain temporally independent. To capture inter-frame relationships essential for action recognition, we perform a temporal aggregation stage that incorporates local inter-frame variation cues and models class-conditioned similarity patterns across frames.

\paragraph{Motion-aware Modulation}
Prior to temporal aggregation, we apply motion-aware modulation to derive local inter-frame variation cues from consecutive CLIP image features.
Specifically, given consecutive frame features $F_{img}^t$ and $F_{img}^{t+1}$, we concatenate them and compute a two-channel transition representation using a $3 \times 3$ convolution $\psi$ as $u^t = \tanh(\psi_{3 \times 3}([F_{img}^t, F_{img}^{t+1}]))$, where $u^t \in \mathbb{R}^{2 \times H \times W}$ denotes the transition representation, and $[\cdot, \cdot]$ represents channel-wise concatenation. The transition representation $u^t$ is projected into a hidden dimension $d_f$ via a $1 \times 1$ convolution $\phi$, resulting in $\gamma^t = \alpha \tanh(\phi(u^t)) \in \mathbb{R}^{H \times W \times d_f}$, with the constant scaling factor for modulation intensity $\alpha = 0.5$.
Subsequently, modulation is applied on the spatially aggregated features $Z_{sa}^t$ as follows:
\begin{equation}
    \begin{aligned}
        \tilde{Z}_{sa}(t,:,:,c,:) \leftarrow (1 + \gamma^t) \odot Z_{sa}(t,:,:,c,:),
        \\
        \text{for}\;\; t=1,\dots,T, \;\; c\in \mathcal{I}_M.
    \end{aligned}
\end{equation}
Here, $\odot$ denotes element-wise multiplication, and the same modulation is applied to each candidate class. In practice, the transition cue estimated from $(t,t+1)$ is applied to frame $t$, and identity modulation is used for the last frame.
Through this operation, $\gamma^t$ helps emphasize representations in dynamically changing regions (\textit{e.g.}, locations where object motion exists), enabling the subsequent Mamba-based temporal aggregation to accumulate motion-related evidence over time.

\paragraph{Mamba-based Temporal Aggregation}
After motion-aware modulation, the similarity volume contains class-conditioned local evidence enriched with adjacent-frame transition cues, but these cues still need to be integrated across the full temporal extent of the video.
For each fixed spatial location $(i,j)$ and sampled class $c$, we regard $\tilde{Z}_{sa}(:,i,j,c,:)$ as a temporal sequence of length $T$ and apply a Mamba~\cite{mamba} block $A_{ta}$:
\begin{align}
    Z_{ta}(:, i, j, c, :) = A_{ta}(\tilde{Z}_{sa}(:, i, j, c, :)).
\end{align}
Mamba serves as a selective state-space temporal operator that performs input-dependent state updates along the temporal sequence. We use a \emph{forward-only scan} over the sampled frame order, allowing transition-aware similarity patterns to be sequentially accumulated while preserving the temporal direction of the video. The scan integrates the modulated similarity patterns across the sampled frames to produce temporally contextualized matching evidence. This design keeps the sequence length bounded by $T$ and processes the $H \times W \times M$ class-conditioned local sequences in parallel, preserving spatio-temporal alignment while efficiently refining temporal similarity patterns.

\subsection{Training Objective}

After temporal aggregation, a linear projection followed by spatio-temporal average pooling produces the video-level class score:
\begin{equation}
    \begin{aligned}
        s(t,i,j,c) = \mathrm{Linear}\!\left(Z_{ta}(t,i,j,c,:)\right),
        \\
        \hat{y}_c = \frac{1}{THW}\sum_{t,i,j} s(t,i,j,c),
        \\
        \quad c \in \mathcal{I}_M.
    \end{aligned}
\end{equation}
The resulting scores are defined over the sampled candidate classes $\mathcal{I}_M$.
During training, the ground-truth class is always included in $\mathcal{I}_M$, and we apply a cross-entropy loss $\mathcal{L}_{agg}$ to these sampled logits using the ground-truth label within the sampled class set.
For settings where class sampling is omitted, the same objective is applied over all action classes.
In addition, we use an auxiliary cross-entropy loss $\mathcal{L}_{cls}$ computed from the global logit $\hat{y}_{cls}$, which is derived from the cosine similarity between $\bar{V}$ and $F_{text}$.
This auxiliary objective retains a global video-text classification signal alongside the local similarity aggregation objective during CLIP adaptation~\cite{vificlip}.
The final training objective is:
\begin{align}
    \mathcal{L} = \mathcal{L}_{agg} + \mathcal{L}_{cls}.
\end{align}

%% file: section/3_exp1.tex
\section{Experiments}
\label{sec:exp}

\subsection{Datasets and Evaluation Protocols}
\label{sec:exp_protocols}
Following prior works, we conduct experiments on five standard action recognition benchmarks: Kinetics-400 (K400)~\cite{k400}, Kinetics-600 (K600)~\cite{k600}, HMDB-51~\cite{hmdb51}, UCF-101~\cite{ucf101}, and Something-Something V2 (SSv2)~\cite{ssv2}.

\noindent{\bf Kinetics-400~\cite{k400} \& Kinetics-600~\cite{k600}} are large-scale datasets sourced from YouTube, containing $\sim$10-second video clips. Kinetics-400 includes approximately 240K training and 20K validation videos across 400 classes. The expanded Kinetics-600 dataset features 600 categories and $\sim$480K total clips. We use Kinetics-600 validation videos for zero-shot transfer evaluation.

\noindent{\bf HMDB51}~\cite{hmdb51} contains 6,869 clips across 51 action classes. We report the average performance over its three standard splits.

\noindent{\bf UCF-101}~\cite{ucf101} consists of 13,320 YouTube-sourced clips covering 101 categories. Similar to HMDB-51, it is evaluated across three standard splits.

\noindent{\bf Something-Something v2 (SSv2)}~\cite{ssv2} comprises 168,913 training and 24,777 validation videos across 174 fine-grained classes. Its strong temporal bias provides a rigorous testbed for evaluating explicit temporal reasoning.

For zero-shot evaluation, the model is pretrained exclusively on Kinetics-400 and evaluated on HMDB-51, UCF-101, and Kinetics-600.
For few-shot evaluation on HMDB-51, UCF-101, and SSv2, the model is directly fine-tuned from pretrained CLIP using $K$ training samples per class, where $K\in\{2,4,8,16\}$.
For base-to-novel evaluation on HMDB-51, UCF-101, and SSv2, classes are divided into base (seen) and novel (unseen) categories. The model is directly fine-tuned from pretrained CLIP using only the base classes of each dataset and evaluated on both base and novel classes. We report the top-1 accuracy for base and novel categories, along with their harmonic mean (HM).

\subsection{Implementation Details}
\label{sec:exp_implementation}
\label{app:setup}
\label{app:AM_details}
We adopt CLIP~\cite{CLIP} with a ViT-B/16 backbone as our base vision-language model for all experiments.
Unless otherwise stated, we uniformly sample 16 frames from each video during training and evaluation. During inference, we adopt a single-view evaluation protocol, using 1 temporal clip and 1 spatial crop per video.

As illustrated in Fig.~\ref{fig:2}, SimVA repeatedly applies the SA-TA structure $N_L$ times; in our implementation, we set $N_L=2$. Each SA module is applied independently for each sampled class and frame, and is implemented with two consecutive Swin Transformer blocks, consisting of window-based and shifted-window self-attention. Each Swin block follows a pre-normalization design with LayerNorm before the self-attention and MLP sublayers.

Each TA module models the temporal evolution of similarity patterns for each class and spatial location. Before the Mamba block, we apply RMSNorm to the temporal sequence at each spatial location. We use a forward Mamba block for temporal aggregation and do not apply additional spatial pooling, preserving patch-level correspondence throughout temporal modeling.

For initialization, the motion-aware modulation module uses Gaussian initialization with a standard deviation of 0.001 for both the $3\times 3$ transition-cue network and the $1\times 1$ modulation projection, with all biases initialized to zero. The Swin Transformer and Mamba blocks follow their default initialization schemes.

\input{table/a_implementation}
Detailed training configurations are provided in Tab.~\ref{tab:hyperparams}. For the zero-shot configuration, SimVA samples $M=100$ out of the 400 available Kinetics-400~\cite{k400} classes per iteration for efficient similarity volume construction. Class sampling is omitted in the few-shot and base-to-novel settings because the target datasets have a small number of classes (51 for HMDB-51, 101 for UCF-101, and 174 for SSv2). All training procedures are conducted using up to 4 NVIDIA RTX A6000 Ada GPUs.
All inference and evaluation procedures across the zero-shot, few-shot, and base-to-novel protocols are conducted on 2 NVIDIA RTX A6000 GPUs.

Following the experimental settings of prior works~\cite{tcclip, bdcclip}, we utilize LLM-generated prompts to enrich the textual representation of action categories.

For the rows marked ``w/ WSE'' in Tab.~\ref{tab:zero_shot}, we apply weight-space ensembling (WSE). The final parameters for both encoders are derived through a linear combination of the pre-trained CLIP checkpoints and the fine-tuned weights, formulated as $\theta_{E} = (1 - \beta) \cdot \theta_{E}^{\text{CLIP}} + \beta \cdot \theta_{E}^{\text{fine-tuned}}$, where $E \in \{ E_{\text{img}}, E_{\text{text}} \}$ and $\beta$ denotes the ensemble ratio. In our experiments, $\beta$ is fixed at 0.8.

\subsection{Main Results}
\noindent{\bf Zero-shot evaluation.}
In Tab.~\ref{tab:zero_shot}, we evaluate zero-shot transferability on HMDB-51, UCF-101, and Kinetics-600.
Without weight-space ensembling (WSE), SimVA achieves the highest mean top-1 accuracy among the compared methods on all three datasets. With WSE, SimVA achieves the highest average top-1 accuracy across the three datasets and remains competitive on K600. More results with 32-frame training are provided in Sec.~\ref{sec:exp_ablation}.

\input{table/zeroshot}

\noindent{\bf Few-shot evaluation.}
In Tab.~\ref{tab:few_shot}, we evaluate the few-shot learning performance of SimVA on HMDB-51, UCF-101 and SSv2. SimVA achieves the best or joint-best accuracy across all $K$-shot settings on the three datasets.
In the 16-shot setting, it improves over the strongest competing methods by 1.3 percentage points on both HMDB-51 and SSv2.
These results demonstrate the effectiveness of SimVA in adapting CLIP to action recognition with limited supervision.

\input{table/fewshot}

\noindent{\bf Base-to-Novel Evaluation.}
In Tab.~\ref{tab:base_to_novel}, we evaluate the base-to-novel generalization performance of SimVA on HMDB-51, UCF-101 and SSv2. SimVA achieves the highest base accuracy across all three datasets and the highest novel accuracy and HM on UCF-101 and SSv2, with HM scores of 88.2\% and 16.5\%, respectively. On HMDB-51, SimVA obtains an HM of 65.3\%, close to the best reported result of 65.5\%. These results indicate that SimVA effectively balances recognition performance on base and novel action categories.

\input{table/base2novel}

\subsection{Computational Cost}
Tab.~\ref{tab:efficiency} compares the computational cost of SimVA with existing CLIP-based OVAR methods. SimVA uses 124.6M parameters and 315 GFLOPs, with a reported throughput of 29. These results indicate a computational cost comparable to the recent CLIP-based methods listed in the table.


\subsection{Ablation Studies}
\label{sec:exp_ablation}
\label{app:frame}
\label{app:videomamba}
To validate the contribution of each component in SimVA, unless otherwise stated, we conduct ablation studies following the zero-shot setting (w/o WSE) and K400 pretraining setup as in Tab.~\ref{tab:zero_shot}.

\input{table/a_cost_component}

\noindent{\bf Effect of individual components.}
Tab.~\ref{tab:component_ablation} analyzes the contribution of each component in SimVA.
The $S$-only (A) variant predicts action classes from the constructed similarity volume without spatial or temporal aggregation.
Adding either spatial (B) or temporal (C) aggregation improves performance on all three datasets, demonstrating the benefit of contextualizing patch-level similarity responses within and across frames.
Combining spatial and temporal aggregation (D) further improves over either component alone, indicating their complementary benefits. 
Adding motion-aware modulation to (D) yields the full model (F), with further gains of 1.4, 0.7, and 0.5 percentage points on HMDB-51, UCF-101, and K600, respectively.
These results support the effectiveness of incorporating local inter-frame variation cues before temporal aggregation.
The full model achieves the highest mean accuracy on each dataset.

\noindent{\bf Temporal aggregator choice: Mamba vs Transformer.}
To evaluate the choice of temporal operator, we compare Mamba with a Transformer-based alternative (E) in Tab.~\ref{tab:component_ablation}. Both variants (E, F) use identical similarity volume construction, class sampling, spatial aggregation, motion-aware modulation, and training objectives, differing only in the temporal aggregation module. The Mamba variant (F) achieves higher mean accuracy on the three datasets in this comparison, supporting its use as the temporal operator in SimVA.

\noindent{\bf Comparison with VideoMamba.}
We compare the proposed structured aggregation with a variant that replaces it with a flattened bidirectional sequence operator inspired by VideoMamba~\cite{VideoMamba}. As shown in Tab.~\ref{tab:videomamba}, SimVA achieves higher mean accuracy on all three datasets, with a 1.1-percentage-point gain in the average top-1 accuracy. In SimVA, spatial aggregation refines each frame before temporal aggregation processes sequences of length $T$ at individual spatial locations. This comparison supports the structured spatial-temporal design within the evaluated setting.

\input{table/a_videomamba}

\noindent{\bf Feature-space vs similarity-space aggregation.}
Tab.~\ref{tab:feature} compares feature aggregation and similarity aggregation using the same aggregation stack.
The feature-space baseline aggregates CLIP image features $F_{\mathrm{img}}$ before computing video-text similarities, whereas SimVA performs aggregation on class-conditioned patch-level similarity representations.
We evaluate both approaches under the zero-shot setting on HMDB-51, UCF-101, and K600, and the 16-shot setting on SSv2.
On the three zero-shot benchmarks, SimVA improves over feature-space aggregation by 2.6, 2.0, and 1.2 percentage points, respectively.
On SSv2 in the 16-shot setting, the corresponding improvement is 2.7 percentage points. Together, these results support the effectiveness of aggregating class-conditioned matching evidence under the evaluated settings.

\input{table/a_feature}

\noindent{\bf Effect of embedding dimension $d_f$.} 
In Table~\ref{tab:dim}, we evaluate the impact of the projected similarity feature dimension $d_f$. As shown, $d_f=64$ achieves the best average performance. Using $d_f = 32$ lowers mean accuracy on all three datasets relative to $d_f=64$. Increasing the dimension to $d_f = 128$ does not provide consistent gains, with only a marginal improvement on K600 within the standard deviation. These results indicate that $d_f=64$ provides a compact and effective similarity projection, so we use it by default.

\input{table/a_dim}

\noindent{\bf Class sampling size $M$.}
In Tab.~\ref{tab:classsampling}, we analyze the influence of the number of sampled candidate classes $M$ during training. The ground-truth class is always included in the sampled subset.
Among the evaluated settings, $M=100$ achieves the highest average accuracy across the three datasets. Reducing $M$ to 50 lowers accuracy on all three datasets, while increasing it to 200 does not improve average performance.
We therefore use $M=100$ as the default setting.

\input{table/a_class}


\noindent{\bf Prompt augmentation.}
We compare LLM-generated prompts with the vanilla CLIP template (e.g., ``A video of \{\}'') in Tab.~\ref{tab:llm}. LLM-generated prompts improve mean accuracy on HMDB-51 and UCF-101, while the vanilla template performs slightly better on K600.

\input{table/a_llm}

\noindent{\bf Robustness to the number of training frames.}
In Kinetics-400~\cite{k400} pre-training based zero-shot action recognition, a 16-frame sequence is widely adopted as the standard configuration~\cite{tcclip, vificlip, xclip,Actionclip}. To comprehensively evaluate how the recent state-of-the-art method, BDC-CLIP~\cite{bdcclip}, handles varying temporal lengths, we investigate its frame-efficiency and report the comparative results in Tab.~\ref{tab:frame_robustness}.
Table~\ref{tab:frame_robustness} shows that BDC-CLIP improves with 32-frame training, whereas SimVA achieves similar results with 16 and 32 frames on UCF-101 and K600 and its highest HMDB-51 accuracy with 16 frames. These results indicate that the 16-frame configuration is effective for SimVA under the evaluated training settings.
\input{table/a_frame}

\subsection{Analysis of Temporal Motion Cues}
\label{app:motion}

To examine SimVA’s ability to exploit temporal motion cues, we evaluate subsets of HMDB-51 and UCF-101 containing classes that are difficult to recognize from a single frame. To reduce subjectivity in class selection, we exclude candidate classes for which single-frame CLIP achieves a per-class accuracy of 60\% or higher, resulting in 10 and 8 classes, respectively. The HMDB-51 subset includes catch, throw, sit, stand, run, walk, clap, wave, flic flac, and somersault. The UCF-101 subset includes Boxing Punching Bag, Boxing Speed Bag, Breast Stroke, Front Crawl, Cricket Bowling, Cricket Shot, Handstand Pushups, and Handstand Walking.

As shown in Tab.~\ref{tab:motion}, SimVA outperforms BDC-CLIP by 4.6 and 4.2 percentage points on the two subsets, exceeding the corresponding gains of 3.8 and 1.5 percentage points on the full class sets. These results suggest that SimVA’s spatio-temporal similarity aggregation is effective for recognizing actions that are difficult to distinguish using static visual cues alone. Additional qualitative results are presented in Fig.~\ref{fig:5}.

\input{table/a_motion}

\subsection{Qualitative Analysis}
\label{ex:vis}
\label{app:vis}

\begin{figure*}[!t]
  \centering
  \includegraphics[width=0.9\textwidth]{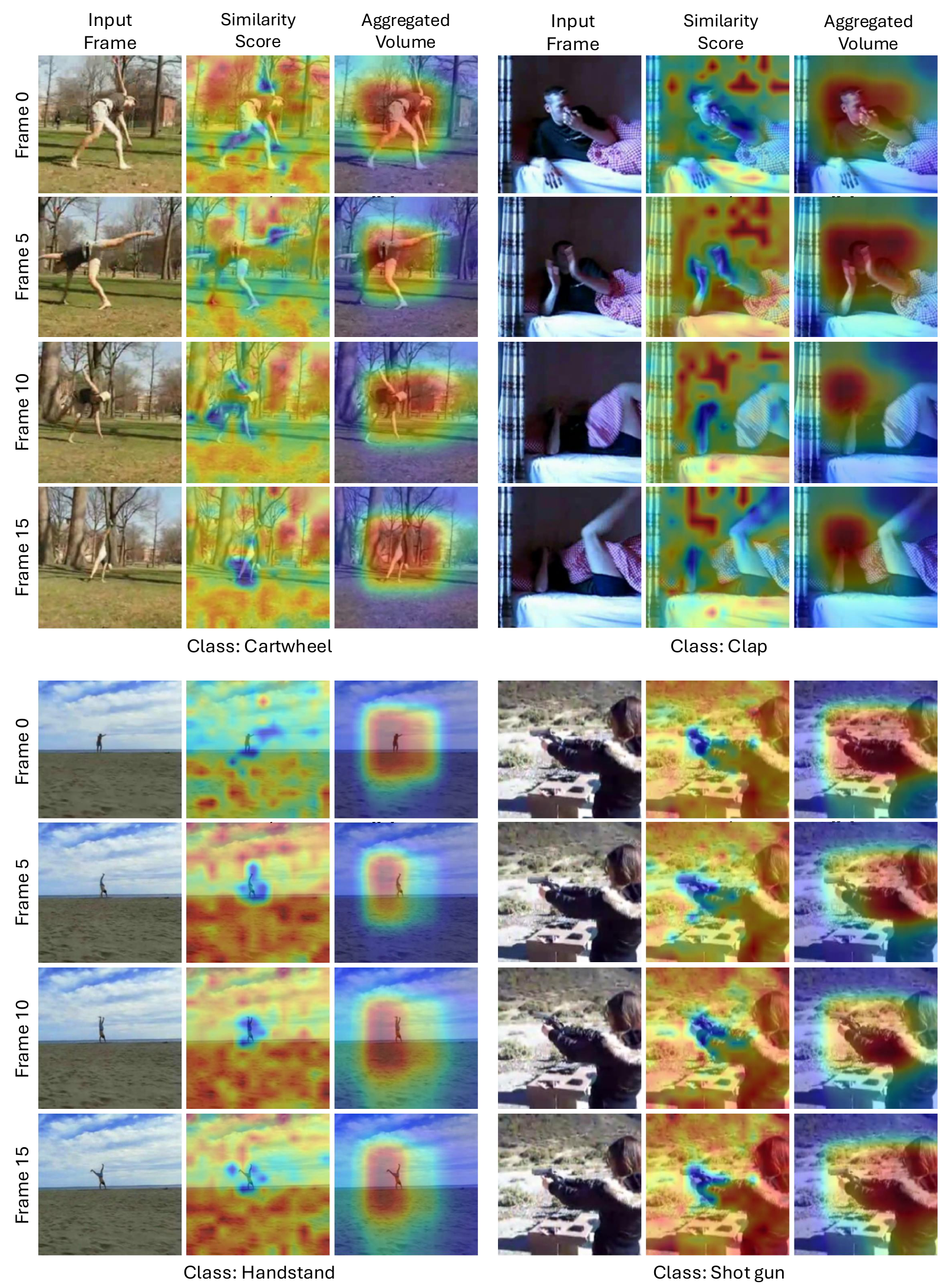}
  \caption{\textbf{Qualitative visualizations of similarity volume aggregation on the HMDB-51 dataset.} For each example, columns from left to right show the input frame, similarity score $S$ (in~\eqref{eq:c}), and aggregated similarity volume. The aggregated volume refines noisy patch-level similarities into more spatially coherent action-related responses across frames.
  }
  \label{fig:appendix_qualitative}
\end{figure*}

\noindent{\bf Similarity volume aggregation.}
We provide qualitative visualizations to further examine the effect of the proposed similarity volume aggregation. For each video sample, we uniformly select four frames and visualize the similarity scores $S$ together with the aggregated similarity volumes produced by SimVA. The similarity score corresponds to the patch-level video-text cosine similarity in Eq.~\eqref{eq:c}, before any aggregation is applied. Since the scores are computed directly from visual patch embeddings and text embeddings, they may exhibit unstable responses in background regions or areas that are not directly related to the target action. In contrast, the aggregated volumes alleviate such scattered activations and produce more coherent responses around action-relevant regions.

As shown in Fig.~\ref{fig:appendix_qualitative}, SimVA refines local video-text matching through spatial and temporal aggregation, rather than directly relying on similarity maps. 
Discriminative visual cues, including human poses, motion-relevant regions, and manipulated objects, are more clearly emphasized after aggregation. 
These examples illustrate how aggregation refines the spatial distribution of class-conditioned matching responses across frames.

\begin{figure*}[t]
  \centering
  \includegraphics[width=\textwidth]{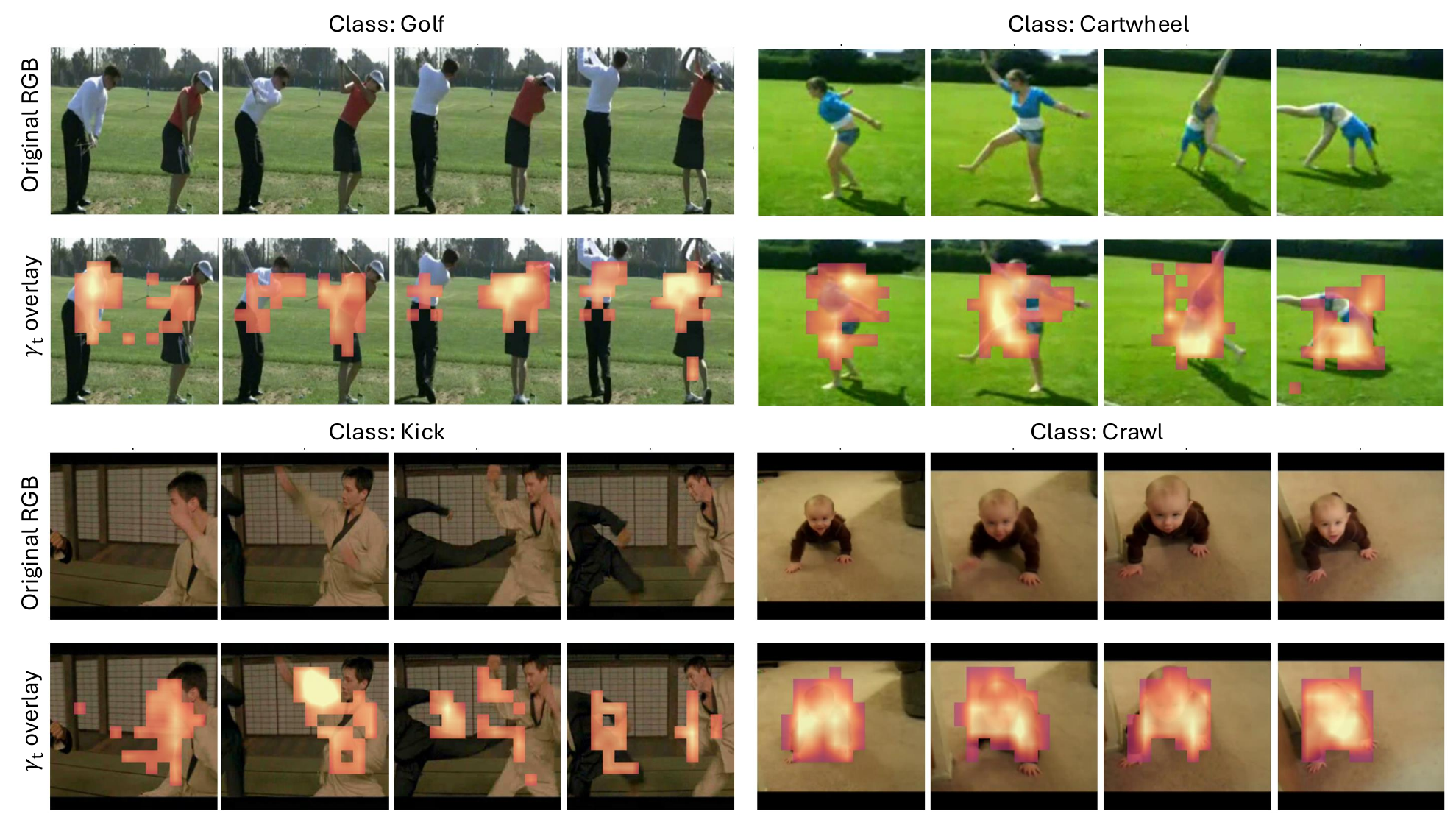}
  \caption{\textbf{Visualization of motion-aware modulation}. For each action example, the upper row shows sampled RGB frames, and the lower row overlays the corresponding modulation responses $\gamma_t$ derived from consecutive CLIP image features. The overlays show spatially varying responses around action-relevant body regions, illustrating how local inter-frame variation cues are incorporated into spatially aggregated similarity representations before temporal aggregation.
  }
  \label{fig:3}
\end{figure*}

\noindent{\bf Motion-aware modulation.}
Fig.~\ref{fig:3} illustrates spatially varying inter-frame responses around action-relevant regions. These examples provide qualitative insight into the variation cues used by motion-aware modulation.

%% file: table/a_implementation.tex
\linespread{1.2}
\begin{table}[t]
\caption{\textbf{Experimental setup.} We summarize the optimization parameters and training schedules utilized across all three evaluation settings: zero-shot, few-shot, and base-to-novel.}
\label{tab:hyperparams}
\centering
\resizebox{\linewidth}{!}{
\begin{tabular}{l|ccc}
\Xhline{0.12em}
\multicolumn{1}{c|}{Configuration}
& Zero-Shot
& Few-Shot
& Base-to-Novel \\
\hline

\multicolumn{4}{l}{\textit{Optimization}} \\

\quad Optimizer
& AdamW & AdamW & AdamW \\

\quad Learning Rate (CLIP Backbone)
& 2e-06 & 2e-06 & 2e-06 \\

\quad Weight Decay
& 0.01 & 0.01 & 0.01 \\

\hline

\multicolumn{4}{l}{\textit{Training Schedule}} \\

\quad Training Frames
& 16 & 16 & 16 \\

\quad Batch Size
& 64 & 32 & 32 \\

\quad Training Epochs
& 5 & 60 & 12 \\

\hline

\multicolumn{4}{l}{\textit{SimVA Specifics}} \\

\quad Similarity Volume Feature Dimension ($d_f$)
& 64 & 64 & 64 \\

\quad Action Class Sampling ($M$)
& 100 & -- & -- \\

\quad Constant Scaling Factor ($\alpha$)
& 0.5 & 0.5 & 0.5 \\

\quad Learning Rate (Aggregator)
& 1e-05 & 1e-04 & 1e-04 \\

\Xhline{0.12em}
\end{tabular}}
\end{table}
\linespread{1}

%% file: table/zeroshot.tex
\linespread{1.4}
\begin{table*}[t]
\caption{\textbf{Zero-shot action recognition} performance on three benchmark datasets, with models pretrained on Kinetics-400. WSE denotes weight-space ensembling, and BEs denote CLIP backbone encoders, which are either frozen (\textbf{F}) or fine-tuned (\textbf{T}). The best and second-best results are highlighted in \sethlcolor{colorFst}\hl{green} and \sethlcolor{colorSnd}\hl{light green}, respectively. ${\dagger}$ indicates results reproduced with official code.}
\label{tab:zero_shot}
\centering
\resizebox{0.75\linewidth}{!}{
\begin{tabular}{l|l|c|cc|cc|c}
\Xhline{0.12em}
& \multicolumn{1}{c|}{Method}
& \multicolumn{1}{c|}{BEs}
& HMDB-51
& UCF-101
& K600 (Top-1)
& K600 (Top-5)
& All (Top-1) \\
\hline

\multirow{9}{*}{\makecell{w/o\\WSE}}
& A5~\cite{a5}
& \textbf{F}
& 44.3 $\pm$ 2.2
& 69.3 $\pm$ 4.2
& 55.8 $\pm$ 0.7
& 81.4 $\pm$ 0.3
& 56.5 \\

& Vita-CLIP~\cite{vitaclip}
& \textbf{F}
& 48.6 $\pm$ 0.6
& 75.0 $\pm$ 0.6
& 67.4 $\pm$ 0.5
& -
& 63.7 \\

& MoTED~\cite{moted}
& \textbf{F}
& 58.2 $\pm$ 1.1
& 78.3 $\pm$ 0.6
& 69.9 $\pm$ 0.5
& -
& 68.8 \\

& ActionCLIP~\cite{Actionclip}
& \textbf{T}
& 49.1 $\pm$ 0.4
& 68.0 $\pm$ 0.9
& 56.1 $\pm$ 0.9
& 83.2 $\pm$ 0.2
& 57.7 \\

& X-CLIP~\cite{xclip}
& \textbf{T}
& 44.6 $\pm$ 5.2
& 72.0 $\pm$ 2.3
& 65.2 $\pm$ 0.4
& 86.1 $\pm$ 0.8
& 60.6 \\

& ViFi-CLIP~\cite{vificlip}
& \textbf{T}
& 52.3 $\pm$ 0.2
& 78.9 $\pm$ 1.1
& 70.7 $\pm$ 0.8
& 92.1 $\pm$ 0.3
& 67.3 \\

& TC-CLIP~\cite{tcclip}
& \textbf{T}
& 56.8 $\pm$ 0.9
& 83.0 $\pm$ 0.6
& 75.4 $\pm$ 0.9
& 94.7 $\pm$ 0.4
& 71.7 \\

& BDC-CLIP~\cite{bdcclip}$^{\dagger}$
& \textbf{T}
& \cellcolor{colorSnd}58.4 $\pm$ 0.6
& \cellcolor{colorSnd}85.0 $\pm$ 0.8
& \cellcolor{colorSnd}75.9 $\pm$ 0.9
& \cellcolor{colorSnd}94.8 $\pm$ 0.3
& \cellcolor{colorSnd}73.1 \\

& SimVA (Ours)
& \textbf{T}
& \cellcolor{colorFst}62.2 $\pm$ 0.2
& \cellcolor{colorFst}86.5 $\pm$ 0.5
& \cellcolor{colorFst}76.5 $\pm$ 1.0
& \cellcolor{colorFst}95.1 $\pm$ 0.2
& \cellcolor{colorFst}75.1 \\

\hline

\multirow{9}{*}{\makecell{w/\\WSE}}
& ActionCLIP~\cite{Actionclip}
& \textbf{T}
& 51.9 $\pm$ 0.5
& 74.2 $\pm$ 1.0
& 67.5 $\pm$ 1.2
& 90.7 $\pm$ 0.1
& 64.5 \\

& ViFi-CLIP~\cite{vificlip}
& \textbf{T}
& 52.2 $\pm$ 0.7
& 81.0 $\pm$ 0.9
& 73.9 $\pm$ 0.5
& 93.3 $\pm$ 0.3
& 69.0 \\

& Open-VCLIP~\cite{openvclip}
& \textbf{T}
& 53.9 $\pm$ 1.2
& 83.4 $\pm$ 1.2
& 73.0 $\pm$ 0.8
& 93.2 $\pm$ 0.1
& 70.1 \\

& MAXI~\cite{maxi}
& \textbf{T}
& 52.3 $\pm$ 0.7
& 78.2 $\pm$ 0.8
& 71.5 $\pm$ 0.8
& 92.5 $\pm$ 0.4
& 67.3 \\

& OST~\cite{ost}
& \textbf{T}
& 55.9 $\pm$ 1.2
& 79.7 $\pm$ 1.1
& 75.1 $\pm$ 0.6
& 94.6 $\pm$ 0.2
& 70.2 \\

& FROSTER~\cite{froster}
& \textbf{T}
& 54.8 $\pm$ 1.3
& 84.8 $\pm$ 1.1
& 74.8 $\pm$ 0.9
& -
& 71.5 \\

& TC-CLIP~\cite{tcclip}
& \textbf{T}
& 56.0 $\pm$ 0.3
& 85.4 $\pm$ 0.8
& 78.1 $\pm$ 1.0
& 95.7 $\pm$ 0.3
& 73.2 \\

& BDC-CLIP~\cite{bdcclip}$^{\dagger}$
& \textbf{T}
& \cellcolor{colorSnd}59.7 $\pm$ 0.7
& \cellcolor{colorSnd}85.9 $\pm$ 0.9
& \cellcolor{colorFst}78.2 $\pm$ 0.9
& \cellcolor{colorSnd}95.7 $\pm$ 0.3
& \cellcolor{colorSnd}74.6 \\

& SimVA (Ours)
& \textbf{T}
& \cellcolor{colorFst}62.1 $\pm$ 0.4
& \cellcolor{colorFst}86.2 $\pm$ 0.6
& \cellcolor{colorSnd}78.1 $\pm$ 1.0
& \cellcolor{colorFst}95.7 $\pm$ 0.2
& \cellcolor{colorFst}75.5 \\

\Xhline{0.12em}
\end{tabular}}
\end{table*}
\linespread{1}

%% file: table/fewshot.tex
\linespread{1.4}
\begin{table*}[t]
\caption{\textbf{Few-shot action recognition} setting with models directly fine-tuned from CLIP. The best and second-best results are highlighted in \sethlcolor{colorFst}\hl{green} and \sethlcolor{colorSnd}\hl{light green}, respectively.}
\label{tab:few_shot}
\centering
\resizebox{0.9\linewidth}{!}{
\begin{tabular}{l|cccc|cccc|cccc}
\Xhline{0.12em}
\multicolumn{1}{c|}{\multirow{2}{*}{Method}}
& \multicolumn{4}{c|}{HMDB-51}
& \multicolumn{4}{c|}{UCF-101}
& \multicolumn{4}{c}{SSv2} \\
\multicolumn{1}{c|}{}
& $K=2$ & $K=4$ & $K=8$ & $K=16$
& $K=2$ & $K=4$ & $K=8$ & $K=16$
& $K=2$ & $K=4$ & $K=8$ & $K=16$ \\
\hline

CLIP~\cite{CLIP}
& 41.9 & 41.9 & 41.9 & 41.9
& 63.6 & 63.6 & 63.6 & 63.6
& 2.7 & 2.7 & 2.7 & 2.7 \\

A5~\cite{a5}
& 39.7 & 50.7 & 56.0 & 62.4
& 71.4 & 79.9 & 85.7 & 89.9
& 4.4 & 5.1 & 6.1 & 9.7 \\

ActionCLIP~\cite{Actionclip}
& 47.5 & 57.9 & 57.3 & 59.1
& 70.6 & 71.5 & 73.0 & 91.4
& 4.1 & 5.8 & 8.4 & 11.1 \\

X-CLIP~\cite{xclip}
& 53.0 & 57.3 & 62.8 & 64.0
& 76.4 & 83.4 & 88.3 & 91.4
& 3.9 & 4.5 & 6.8 & 10.0 \\

ViFi-CLIP~\cite{vificlip}
& 57.2 & 62.7 & 64.5 & 66.8
& 80.7 & 85.1 & 90.0 & 92.7
& 6.2 & 7.4 & 8.5 & 12.4 \\

OST~\cite{ost}
& 59.1 & 62.9 & 64.9 & 68.2
& 82.5 & 87.5 & 91.7 & 93.9
& 7.0 & 7.7 & 8.9 & 12.2 \\

TC-CLIP~\cite{tcclip}
& 58.6
& \cellcolor{colorSnd}63.3
& \cellcolor{colorSnd}65.5
& \cellcolor{colorSnd}68.8
& \cellcolor{colorSnd}86.8
& \cellcolor{colorSnd}90.1
& 92.0
& 94.3
& \cellcolor{colorSnd}7.3
& \cellcolor{colorSnd}8.6
& 9.3
& 14.0 \\

BDC-CLIP~\cite{bdcclip}$^{\dagger}$
& \cellcolor{colorSnd}59.2
& 63.1
& 65.3
& 67.5
& 85.6
& 89.3
& 92.0
& \cellcolor{colorSnd}94.4
& 6.4
& 8.3
& \cellcolor{colorSnd}9.6
& \cellcolor{colorSnd}14.8 \\

SimVA (Ours)
& \cellcolor{colorFst}60.7
& \cellcolor{colorFst}64.0
& \cellcolor{colorFst}67.5
& \cellcolor{colorFst}70.1
& \cellcolor{colorFst}87.0
& \cellcolor{colorFst}90.5
& \cellcolor{colorFst}92.4
& \cellcolor{colorFst}94.4
& \cellcolor{colorFst}7.3
& \cellcolor{colorFst}9.0
& \cellcolor{colorFst}10.3
& \cellcolor{colorFst}16.1 \\

\Xhline{0.12em}
\end{tabular}}
\end{table*}
\linespread{1}

%% file: table/base2novel.tex
\linespread{1.4}
\begin{table*}[t]
\centering

\begin{minipage}[t]{0.60\textwidth}
\vspace{0pt}
\caption{\textbf{Base-to-novel generalization} setting with models directly fine-tuned from CLIP. HM denotes harmonic mean. The best and second-best results are highlighted in \sethlcolor{colorFst}\hl{green} and \sethlcolor{colorSnd}\hl{light green}, respectively.}
\label{tab:base_to_novel}
\end{minipage}
\hfill
\begin{minipage}[t]{0.38\textwidth}
\vspace{0pt}
\caption{\textbf{Computational cost}. TP denotes throughput per view measured on an A6000 Ada GPU.}
\label{tab:efficiency}
\end{minipage}

\par\vspace{2pt}

\begin{minipage}[t]{0.60\textwidth}
\vspace{0pt}
\centering
\resizebox{\linewidth}{!}{
\begin{tabular}{l|ccc|ccc|ccc}
\Xhline{0.12em}
\multicolumn{1}{c|}{\multirow{2}{*}{Method}}
& \multicolumn{3}{c|}{HMDB-51}
& \multicolumn{3}{c|}{UCF-101}
& \multicolumn{3}{c}{SSv2} \\
\multicolumn{1}{c|}{}
& Base & Novel & HM
& Base & Novel & HM
& Base & Novel & HM \\
\hline

CLIP~\cite{CLIP}
& 53.3 & 46.8 & 49.8
& 78.5 & 63.6 & 70.3
& 4.9 & 5.3 & 5.1 \\

A5~\cite{a5}
& 46.2 & 16.0 & 23.8
& 90.5 & 40.4 & 55.8
& 8.3 & 5.3 & 6.5 \\

ActionCLIP~\cite{Actionclip}
& 69.1 & 37.3 & 48.5
& 90.1 & 58.1 & 70.7
& 13.3 & 10.1 & 11.5 \\

X-CLIP~\cite{xclip}
& 69.4 & 45.5 & 55.0
& 89.9 & 58.9 & 71.2
& 8.5 & 6.6 & 7.4 \\

ViFi-CLIP~\cite{vificlip}
& 73.8 & 53.3 & 61.9
& 92.9 & 67.7 & 78.3
& 16.2 & 12.1 & 13.9 \\

TC-CLIP~\cite{tcclip}
& 73.3
& \cellcolor{colorFst}59.1
& \cellcolor{colorFst}65.5
& \cellcolor{colorSnd}95.4
& \cellcolor{colorSnd}81.6
& \cellcolor{colorSnd}88.0
& \cellcolor{colorSnd}17.5
& 13.4
& 15.2 \\

BDC-CLIP~\cite{bdcclip}$^{\dagger}$
& \cellcolor{colorSnd}74.5
& 55.0
& 63.3
& 94.9
& 78.8
& 86.1
& 16.2
& \cellcolor{colorSnd}14.9
& \cellcolor{colorSnd}15.5 \\

SimVA (Ours)
& \cellcolor{colorFst}75.4
& \cellcolor{colorSnd}57.6
& \cellcolor{colorSnd}65.3
& \cellcolor{colorFst}95.5
& \cellcolor{colorFst}82.0
& \cellcolor{colorFst}88.2
& \cellcolor{colorFst}18.0
& \cellcolor{colorFst}15.3
& \cellcolor{colorFst}16.5 \\

\Xhline{0.12em}
\end{tabular}}
\end{minipage}
\hfill
\begin{minipage}[t]{0.38\textwidth}
\vspace{0pt}
\centering
\resizebox{\linewidth}{!}{
\begin{tabular}{l|ccc}
\Xhline{0.12em}
\multicolumn{1}{c|}{Method}
& Params (M)
& GFLOPs
& TP \\
\hline

ActionCLIP~\cite{Actionclip}
& 143.7 & 567 & 23 \\

X-CLIP~\cite{xclip}
& 169.7 & 288 & 40 \\

Vita-CLIP~\cite{vitaclip}
& 161.8 & 307 & 34 \\

ViFi-CLIP~\cite{vificlip}
& 124.3 & 285 & 43 \\

TC-CLIP~\cite{tcclip}
& 127.5 & 304 & 27 \\

BDC-CLIP~\cite{bdcclip}
& 126.9 & 316 & 30 \\

SimVA (Ours)
& 124.6 & 315 & 29 \\

\Xhline{0.12em}
\end{tabular}}
\end{minipage}

\end{table*}
\linespread{1}

%% file: table/a_cost_component.tex
\linespread{1.4}
\begin{table*}[t]
\caption{\textbf{Component-wise ablation.} Rep. denotes the input representation used for aggregation, either image features $F_{\mathrm{img}}$ or the spatio-temporal similarity volume $S$. $A_{\mathrm{sa}}$, $A_{\mathrm{ta}}$, and M.M. denote spatial aggregation, temporal aggregation, and motion-aware modulation, respectively.}
\label{tab:component_ablation}
\centering
\resizebox{0.8\linewidth}{!}{
\begin{tabular}{c|l|c|ccc|ccc}
\Xhline{0.12em}
\multicolumn{1}{c|}{Type}
& \multicolumn{1}{c|}{Case}
& \multicolumn{1}{c|}{Rep.}
& $A_{\mathrm{sa}}$
& $A_{\mathrm{ta}}$
& M.M.
& HMDB
& UCF
& K600 \\
\hline

(A)
& $S$ only
& $S$
& --
& --
& --
& 57.5 $\pm$ 0.8
& 82.7 $\pm$ 1.1
& 73.5 $\pm$ 1.1 \\

(B)
& $S + A_{\mathrm{sa}}$
& $S$
& \checkmark
& --
& --
& 59.1 $\pm$ 0.3
& 85.3 $\pm$ 0.3
& 75.6 $\pm$ 1.2 \\

(C)
& $S + A_{\mathrm{ta}}$
& $S$
& --
& Mamba
& --
& 59.3 $\pm$ 0.4
& 84.5 $\pm$ 0.3
& 75.4 $\pm$ 1.2 \\

(D)
& w/o M.M.
& $S$
& \checkmark
& Mamba
& --
& 60.8 $\pm$ 0.6
& 85.8 $\pm$ 0.2
& 76.0 $\pm$ 1.0 \\

\hline

(E)
& SimVA w/ Transformer
& $S$
& \checkmark
& Transformer
& \checkmark
& 60.8 $\pm$ 0.5
& 85.7 $\pm$ 0.7
& 75.9 $\pm$ 0.9 \\

(F)
& SimVA (Ours)
& $S$
& \checkmark
& Mamba
& \checkmark
& 62.2 $\pm$ 0.2
& 86.5 $\pm$ 0.5
& 76.5 $\pm$ 1.0 \\

\Xhline{0.12em}
\end{tabular}}
\end{table*}
\linespread{1}

%% file: table/a_videomamba.tex
\linespread{1.2}
\begin{table}[!ht]
\caption{\textbf{Comparison with VideoMamba.}}
\label{tab:videomamba}
\centering
\resizebox{\linewidth}{!}{
\begin{tabular}{l|cccc}
\Xhline{0.12em}
\multicolumn{1}{c|}{Method}
& HMDB
& UCF
& K600
& All ($\Delta$) \\
\hline

VideoMamba~\cite{VideoMamba}
& 60.5 $\pm$ 0.4
& 85.7 $\pm$ 0.4
& 75.9 $\pm$ 0.2
& 74.0 \\

SimVA (Ours)
& 62.2 $\pm$ 0.2
& 86.5 $\pm$ 0.5
& 76.5 $\pm$ 1.0
& 75.1 ($+$1.1) \\

\Xhline{0.12em}
\end{tabular}}
\end{table}
\linespread{1}

%% file: table/a_feature.tex
\linespread{1.2}
\begin{table}[!ht]
\caption{\textbf{Comparison of feature-space and similarity-space aggregation using the same aggregation stack.} We report top-1 accuracy (\%) under the zero-shot setting on HMDB-51, UCF-101, and K600, and the few-shot setting on SSv2.}
\label{tab:feature}
\centering
\resizebox{\linewidth}{!}{
\begin{tabular}{l|cccc}
\Xhline{0.12em}
\multicolumn{1}{c|}{Method}
& HMDB
& UCF
& K600
& SSv2 \\
\hline

Feature Aggregation
& 59.6 $\pm$ 0.4
& 84.5 $\pm$ 1.0
& 75.3 $\pm$ 0.4
& 13.4\\

SimVA (Ours)
& 62.2 $\pm$ 0.2
& 86.5 $\pm$ 0.5
& 76.5 $\pm$ 1.0
& 16.1 \\

\Xhline{0.12em}
\end{tabular}}
\end{table}
\linespread{1}

%% file: table/a_dim.tex
\linespread{1.4}
\begin{table}[!ht]
\caption{\textbf{Similarity volume dimension}.}
\label{tab:dim}
\centering
\resizebox{0.7\linewidth}{!}{
\begin{tabular}{c|ccc}
\Xhline{0.12em}
$d_f$ & HMDB & UCF & K600 \\
\hline

32
& 60.4 $\pm$ 0.3
& 85.9 $\pm$ 0.7
& 75.7 $\pm$ 1.4 \\

64
& 62.2 $\pm$ 0.2
& 86.5 $\pm$ 0.5
& 76.5 $\pm$ 1.0 \\

128
& 60.7 $\pm$ 0.2
& 86.0 $\pm$ 0.7
& 76.6 $\pm$ 0.8 \\

\Xhline{0.12em}
\end{tabular}}
\end{table}
\linespread{1}

%% file: table/a_class.tex
\linespread{1.4}
\begin{table}[!ht]
\caption{\textbf{Class sampling $M$}.}
\label{tab:classsampling}
\centering
\resizebox{0.7\linewidth}{!}{
\begin{tabular}{c|ccc}
\Xhline{0.12em}
$M$ & HMDB & UCF & K600 \\
\hline

50
& 60.1 $\pm$ 0.2
& 85.6 $\pm$ 0.7
& 76.3 $\pm$ 1.1 \\

100
& 62.2 $\pm$ 0.2
& 86.5 $\pm$ 0.5
& 76.5 $\pm$ 1.0 \\

200
& 61.0 $\pm$ 0.1
& 86.3 $\pm$ 0.1
& 76.4 $\pm$ 1.0 \\

\Xhline{0.12em}
\end{tabular}}
\end{table}
\linespread{1}

%% file: table/a_llm.tex
\linespread{1.2}
\begin{table}[!ht]
\caption{\textbf{Influence of Text Augmentation on zero-shot action recognition.}}
\label{tab:llm}
\centering
\resizebox{0.7\linewidth}{!}{
\begin{tabular}{l|ccc}
\Xhline{0.12em}
\multicolumn{1}{c|}{Method}
& HMDB
& UCF
& K600 \\
\hline

Vanilla
& 61.3 $\pm$ 0.4
& 85.7 $\pm$ 0.4
& 76.8 $\pm$ 0.2 \\

LLM
& 62.2 $\pm$ 0.2
& 86.5 $\pm$ 0.5
& 76.5 $\pm$ 1.0 \\

\Xhline{0.12em}
\end{tabular}}
\end{table}
\linespread{1}

%% file: table/a_frame.tex
\linespread{1.2}
\begin{table}[!ht]
\caption{Comparison of zero-shot action recognition performance between BDC-CLIP and our method under 16-frame and 32-frame training configurations. ${\dagger}$ indicates results reproduced with official code.}
\label{tab:frame_robustness}
\centering
\resizebox{\linewidth}{!}{
\begin{tabular}{l|cccc}
\Xhline{0.12em}
\multicolumn{1}{c|}{Method}
& HMDB-51
& UCF-101
& K600 Top-1
& K600 Top-5 \\
\hline

\multicolumn{5}{l}{\textit{Training with 16 frames}} \\
\hline

BDC-CLIP$^{\dagger}$~\cite{bdcclip}
& 58.4 $\pm$ 0.6
& 85.0 $\pm$ 0.8
& 75.9 $\pm$ 0.9
& 94.8 $\pm$ 0.3 \\

SimVA (Ours)
& 62.2 $\pm$ 0.2
& 86.5 $\pm$ 0.5
& 76.5 $\pm$ 1.0
& 95.1 $\pm$ 0.2 \\

\hline

\multicolumn{5}{l}{\textit{Training with 32 frames}} \\
\hline

BDC-CLIP~\cite{bdcclip}
& 59.4 $\pm$ 0.3
& 85.9 $\pm$ 0.9
& 76.5 $\pm$ 0.8
& 95.0 $\pm$ 0.3 \\

SimVA (Ours)
& 61.6 $\pm$ 0.5
& 86.5 $\pm$ 0.4
& 76.5 $\pm$ 1.0
& 95.0 $\pm$ 0.3 \\

\Xhline{0.12em}
\end{tabular}}
\end{table}
\linespread{1}

%% file: table/a_motion.tex
\linespread{1.2}
\begin{table}[!ht]
\caption{\textbf{Accuracy (\%) on class subsets that are difficult to recognize from a single frame (m) and the full class sets (f).}}
\label{tab:motion}
\centering
\resizebox{\linewidth}{!}{
\begin{tabular}{l|cccc}
\Xhline{0.12em}
\multicolumn{1}{c|}{Method}
& HMDB (m)
& HMDB (f)
& UCF (m)
& UCF (f)\\
\hline

TC-CLIP ~\cite{tcclip}
& 26.0 $\pm$ 0.8
& 56.8 $\pm$ 0.9
& 56.3 $\pm$ 2.0
& 83.0 $\pm$ 0.6 \\

BDC-CLIP ~\cite{bdcclip}
& 31.2 $\pm$ 1.1
& 58.4 $\pm$ 0.6
& 67.5 $\pm$ 5.0
& 85.0 $\pm$ 0.8 \\

SimVA (Ours)
& 35.8 $\pm$ 1.8
& 62.2 $\pm$ 0.2
& 71.7 $\pm$ 2.1
& 86.5 $\pm$ 0.5 \\

\Xhline{0.12em}
\end{tabular}}
\end{table}
\linespread{1}

%% file: section/4_related.tex
\section{Related Works}
\subsection{Cost Volume}
Cost volume has been widely adopted in dense correspondence tasks such as optical flow, stereo matching, and visual correspondence. Early works such as RAFT~\cite{raft} construct dense all-pairs correlation volumes and iteratively refine them to estimate motion fields. Transformer-based approaches, including CATS~\cite{cats} and CostFormer~\cite{costformer}, further enhance cost aggregation through attention-based refinement. Extending this paradigm to open-vocabulary semantic segmentation, Cat-Seg~\cite{Catseg} incorporates cost aggregation within a vision–language framework to preserve dense patch-level correspondences. Building on cost aggregation for spatial correspondence and semantic matching, our work extends the refinement of visual-text matching evidence to the spatio-temporal domain. Specifically, we construct a 4D similarity volume between video patches and textual embeddings and design a structured spatial–temporal aggregation architecture to model both intra-frame spatial consistency and inter-frame motion dynamics for open-vocabulary action recognition.

\subsection{Open-Vocabulary Action Recognition (OVAR)}
OVAR extends CLIP~\cite{CLIP} to the video domain by modeling temporal dynamics while preserving vision–language alignment. Early approaches such as ActionCLIP~\cite{Actionclip} and X-CLIP~\cite{xclip} introduced temporal modeling modules on top of CLIP, whereas ST-Adapter~\cite{stadapter} and AIM~\cite{aim} proposed lightweight adaptation strategies for efficient image-to-video transfer. ViFi-CLIP~\cite{vificlip} further demonstrated that full fine-tuning of CLIP can effectively reduce the domain gap between images and videos. Recent works have focused on improving video–text alignment. VITA-CLIP~\cite{vitaclip} and prompt-based methods adapt visual–text representations via multimodal prompting, while Open-VCLIP~\cite{openvclip} transforms CLIP into an open-vocabulary video model through weight interpolation. TC-CLIP~\cite{tcclip} contextualizes visual tokens across frames, while BDC-CLIP~\cite{bdcclip} models dependencies between visual and textual token representations through Brownian Distance Covariance. In contrast, we move beyond feature-level aggregation by operating directly in the similarity space, preserving dense patch-level correspondences between video and text for spatio-temporal reasoning.

%% file: section/5_conclusion.tex
\section{Conclusion}
\label{sec:conclusion}
In this paper, we propose SimVA, a similarity volume aggregation framework for open-vocabulary action recognition. By constructing dense patch-level similarities between video and text, SimVA retains class-conditioned matching patterns for refinement before video-level pooling. The framework combines training-time class sampling with structured spatial-temporal aggregation and motion-aware modulation. Experiments across zero-shot, few-shot, and base-to-novel settings show competitive performance, with gains over strong baselines in multiple evaluated conditions. These results show that similarity-space aggregation is an effective direction for adapting vision-language models to video action recognition.